\documentclass[10pt,twocolumn,letterpaper]{article}

\usepackage{iccv}
\usepackage{times}
\usepackage{epsfig}
\usepackage{graphicx}
\usepackage{amsmath}
\usepackage{amssymb}

\usepackage[breaklinks=true,bookmarks=false]{hyperref}

\usepackage{enumitem}

\usepackage{mathtools}
\usepackage{multicol}
\usepackage{multirow}
\usepackage[table]{xcolor}
\usepackage{subcaption} 
\usepackage{caption} 
\usepackage{authblk}
\iccvfinalcopy
\def\iccvPaperID{38} %

\ificcvfinal\fi

\begin{document}

\title{Identifying Systematic Errors in Object Detectors with the SCROD Pipeline}
\author[1,2]{Valentyn Boreiko}
\author[2]{Matthias Hein}
\author[1]{Jan Hendrik Metzen}
\affil[1]{Bosch Center for Artificial Intelligence, Robert Bosch GmbH}
\affil[2]{University of Tübingen}
\maketitle
\ificcvfinal\thispagestyle{empty}\fi

\begin{abstract}
   The identification and removal of systematic errors in object detectors can be a prerequisite for their deployment in safety-critical applications like automated driving and robotics.  
   Such systematic errors can for instance occur under very specific object poses (location, scale, orientation), object colors/textures, and backgrounds.
   Real images alone are unlikely to cover all relevant combinations. We overcome this limitation by generating synthetic images with fine-granular control. 
   While generating synthetic images with physical simulators and hand-designed 3D assets allows fine-grained control over generated images, this approach is resource-intensive and has limited scalability. In contrast, using generative models is more scalable but less reliable in terms of fine-grained control. In this paper, we propose a novel framework that combines the strengths of both approaches. Our meticulously designed pipeline along with custom models enables us to generate street scenes with fine-grained control in a fully automated and scalable manner. Moreover, our framework introduces an evaluation setting that can serve as a benchmark for similar pipelines. This evaluation setting will contribute to advancing the field and promoting standardized testing procedures.
\end{abstract}

\section{Introduction}
\begin{figure}[t]
  \centering
    \centering
    \includegraphics[width=\linewidth]{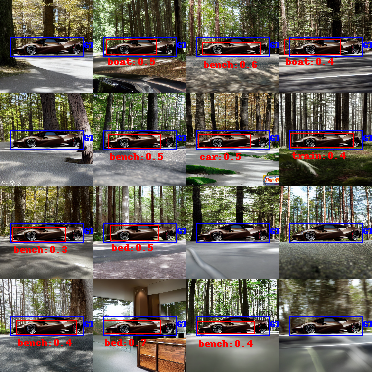}
    \caption{\textbf{Systematic errors detected with our SCROD pipeline into wrong classes \textit{boat, bench, bed, and train} or False Negatives}. The prompt used is \textit{``brown sports car is driving in forest, , in the morning, bright, idyllic, insanely detailed.''}. Here, the image is downscaled by a factor of $5.5$ before inputting it into the object detector, the rotation angle is $-20^{\circ}$, and images are generated with $16$ different seeds. Only in one case, it is detected as correct class \textbf{car}. The used object detector is \textit{FasterRCNN\_ResNet50\_FPN} from torchvision \cite{torchvision2016}.}
  \label{figure:teaser}
\end{figure}

Deep learning has significantly improved performance in many computer vision domains \cite{liu_deep_2020,9356353}. However, models can display subpar performance on narrow but semantically coherent subgroups of the data. %
This can happen due to spurious features \cite{singla2022salient,yannic2023spurious}, associations that a computer vision model has picked up when utilizing shortcut learning \cite{geirhos2020shortcut,unbiased_look_at_dataset_bias,li_2021_discover,Li_2022_discover}. While such shortcuts can lead to higher accuracy on the in-distribution data, they often fail on out-out-distribution data \cite{pmlr-v139-zhou21g} and in the long-tail of the data distribution. 

This non-homogeneous performance is problematic for applications such as automated driving, where models should work well on all subgroups of the data \cite{Blank_Hueger_Mock_Stauner_2022}. Therefore, recently there has been an increased interest in identifying subgroups of the data on which a computer vision model has subpar performance, which we call \textit{systematic errors} \cite{eyuboglu_domino:_2022,jain_distilling_2022,Metzen_Systematic_Errors,Tong_Mass_Producing,wiles2022discovering}.
While systematic errors have been studied for image classification \cite{eyuboglu_domino:_2022,jain_distilling_2022,Metzen_Systematic_Errors,wiles2022discovering}, multi-modal generative models \cite{Tong_Mass_Producing} and image captioning \cite{gao2022adaptive}, there is limited research for their discovery for object detectors, relying on a human-in-the-loop \cite{gao2022adaptive}. 

\begin{figure}
  \begin{minipage}[b]{0.5\linewidth}
    \centering
    \includegraphics[width=\linewidth]{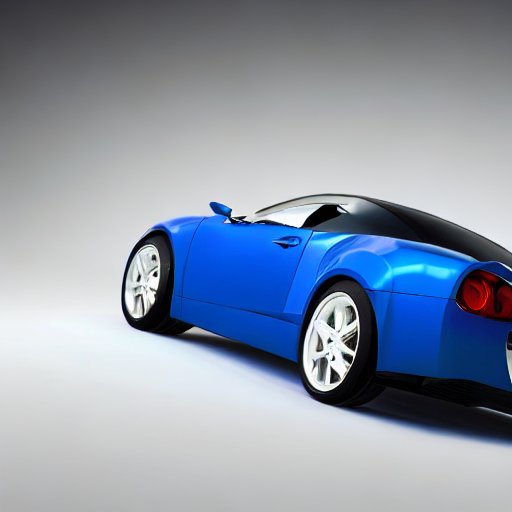}
    \label{fig:image1}
  \end{minipage}%
  \begin{minipage}[b]{0.5\linewidth}
    \centering
    \includegraphics[width=\linewidth]{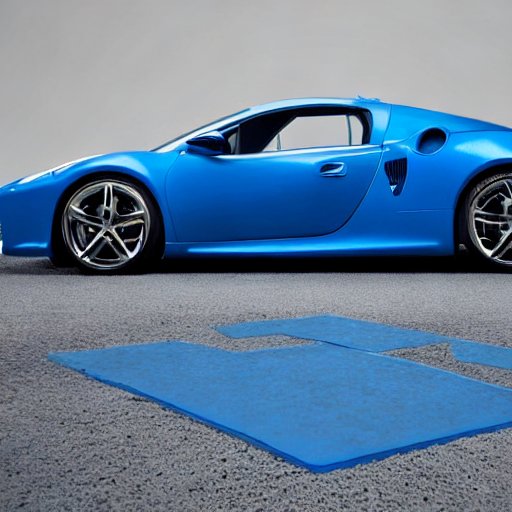}
    \label{fig:image2}
  \end{minipage}

  \vspace{-1em}

  \begin{minipage}[b]{0.5\linewidth}
    \centering
    \includegraphics[width=\linewidth]{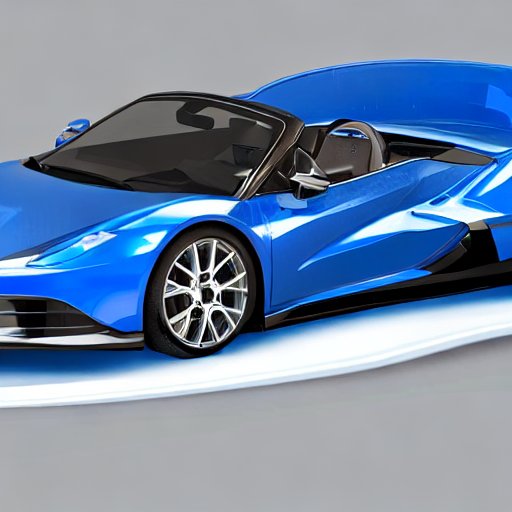}
    \label{fig:image3}
  \end{minipage}%
  \begin{minipage}[b]{0.5\linewidth}
    \centering
    \includegraphics[width=\linewidth]{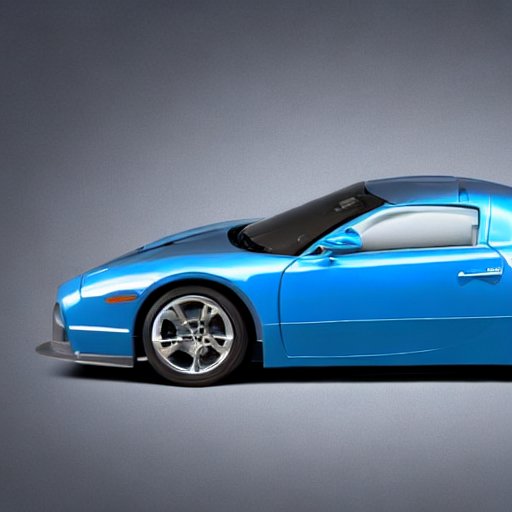}
    \label{fig:image4}
  \end{minipage}
  \caption{\textbf{Single generative models, such as a  Stable Diffusion v1.5 (SD-v1.5) \cite{rombach2021highresolution}, do not allow to reliably generate images with fine-grained control, \textit{while our SCROD pipeline can (see Fig.~\ref{fig:highest_error_rate_fasterRCNN})}}. In this figure, we show images generated by SD-v1.5 for $4$ different seeds and the prompt \textit{``blue sports car rotated by 50 degrees around the X axis from the side view on a grey background in the center of the image''}. In general, orientation is neither consistent nor correct,  the object is cropped, and the color of the background and the main object is confused (upper right image, ``attribute binding'').}
  \label{fig:failure_stable_diffusion}
\end{figure}

Real-world data is unlikely to cover all relevant subgroups in the tail of the data distribution. A promising alternative is to evaluate object detectors on synthetic data instead. Methods for synthesizing data can be roughly split into two categories: i) methods that rely on using physical simulators and hand-designed 3D assets \cite{leclerc2021three,Resnick2021CBERT_Wokrshop}, which is labor-intensive and not scalable; ii) methods that rely on a single generative model, such as the Stable Diffusion v1.5 (SD-v1.5) \cite{rombach2021highresolution}, where geometric properties such object location, scale, or orientation are not easily controlled by textual prompts, as can be seen in the Fig.~\ref{fig:failure_stable_diffusion}. Moreover, the well-known problem of \textit{attribute binding} makes also object color/texture, and background control unreliable.

We address this challenge by combining several generative models (externally pre-trained as well as custom fine-tuned ones) in our novel pipeline (see Fig.~\ref{fig:pipeleine}). While we focus on the application of automated driving and thus on car objects, our method can be extended to other types of objects that are well represented in the distribution of the training data of the generative models we use in our pipeline. With this pipeline, we can identify systematic errors of object detectors (Fig.~\ref{figure:teaser}).

Our contributions are as follows:
\begin{itemize}[noitemsep,topsep=0pt,parsep=0pt,partopsep=0pt]
    \item In Section \ref{sec:pipeline}, we propose a \emph{novel pipeline for street scene synthesis}, allowing fine-grained control over attributes such as object location, scale, orientation,  color, type, as well as scene background.
    \item We propose a custom model for \emph{outpainting}, which is based on finetuning an inpainting model for street-scene outpainting, that we use in our street scene synthesis pipeline for background generation.
    \item In Section \ref{sec:evaluation}, we conduct an evaluation of the object detectors using our pipeline in two settings: one, where the background is a plain color and a second one where we show how our findings extend to more realistic (outpainted) backgrounds, where the background is generated conditioned on the text prompt. There we show some concrete systematic errors of the best object detectors from torchvision \cite{torchvision2016}.
\end{itemize}

\section{Related work}

\textbf{Controlled image generation.}
Controlled image generation can be achieved either by training generative models such as Stable Diffusion models from scratch with some guidance, such as inpainting guidance \cite{rombach2021highresolution} or using methods for finetuning \cite{zhang2023adding,hu2022lora,mou2023t2iadapter,li2023gligen,xue2023freestylenet,cheng2023layoutdiffuse}, with guidance such as inpainting, Canny Edges, Depths Maps, Segmentation Maps, bounding boxes with target classes and many more. None of them can however offer a reliable fine-grained control over all the attributes at the same time that are useful for object detection: \textit{location, scale, orientation, object color, object type, and background}. 

\textbf{Systematic error identification with subgroup annotations.}
This research direction mainly relies on creating reliable datasets, that can label as many attributes (or subgroups) as possible. Some examples are: DeepFashion2 \cite{DeepFashion2} (attributes such as occlusions, segmentation, viewpoint, style, and category name are labeled), ImageNet-X \cite{Idrissi2022ImageNetX} (attributes such as pose and background are labeled), \textbf{WEDGE} \cite{Marathe_2023_CVPR}, \textbf{DAWN} \cite{MouradDAWN}, \textbf{nuScenes} \cite{nuscenes} (attributes such as weather condition). While it is the most reliable way to test object detectors, it is not scalable as covering all possible combinations of relevant attributes is not feasible in general. Note moreover, that only the datasets in bold are related to the application of automated driving. Thus, this research offers limited utility for the systematic error identification of object detectors.
Additionally, methods such as \cite{leclerc2021three,Resnick2021CBERT_Wokrshop} also offer subgroup annotations, but they rely on hand-made 3D assets or physical simulators. 

\textbf{Systematic error identification without subgroup annotations.}
Previous methods \cite{eyuboglu_domino:_2022,jain_distilling_2022,Metzen_Systematic_Errors,Tong_Mass_Producing,wiles2022discovering} have predominantly focused on the systematic errors of either classifiers or multi-modal models such as CLIP \cite{pmlr-v139-radford21a}, and only a few, such as AdaVision \cite{gao2022adaptive}, have done systematic error identification without subgroup annotations for object detection models. AdaVision, however, proposes a search over existing images with human-in-the-loop. It thus does not allow fine-grained control over the object attributes, as it is challenging to collect a dataset that encompasses all possible combinations of attributes, retrieve the relevant ones, and search through them automatically.

\textbf{Novel View Synthesis}.
Control over the orientation of the object is important for automatically testing object detectors. Currently, this can be achieved either by using 3D assets and physical simulators as mentioned above or by using a generative model, such as Stable Diffusion, and fine-tuning it to predict novel views from a single view such as in Zero-1-to-3 \cite{liu2023zero1to3}. This latter approach is promising for our pipeline and it can be further improved by fine-tuning on a bigger dataset \cite{deitke_objaverse_xl}.

\section{Segment Control Rotate Outpaint Detect (SCROD) Pipeline}\label{sec:pipeline}

To identify systematic errors of object detectors, we require fine-grained control of the object properties such as pose. Using a single generative model, such as SD-v1.5, as has been done by Metzen et al.\,\cite{Metzen_Systematic_Errors}, does not allow such fine-grained control of details of the generated object to the degree which is necessary for the testing of object detectors, as can be seen in Fig.~\ref{fig:failure_stable_diffusion}. 

\subsection{Workflow of SCROD}
\begin{figure*}[hbt!]
  \centering
  \begin{subfigure}{1.0\textwidth}
    \centering
    \includegraphics[width=\linewidth]{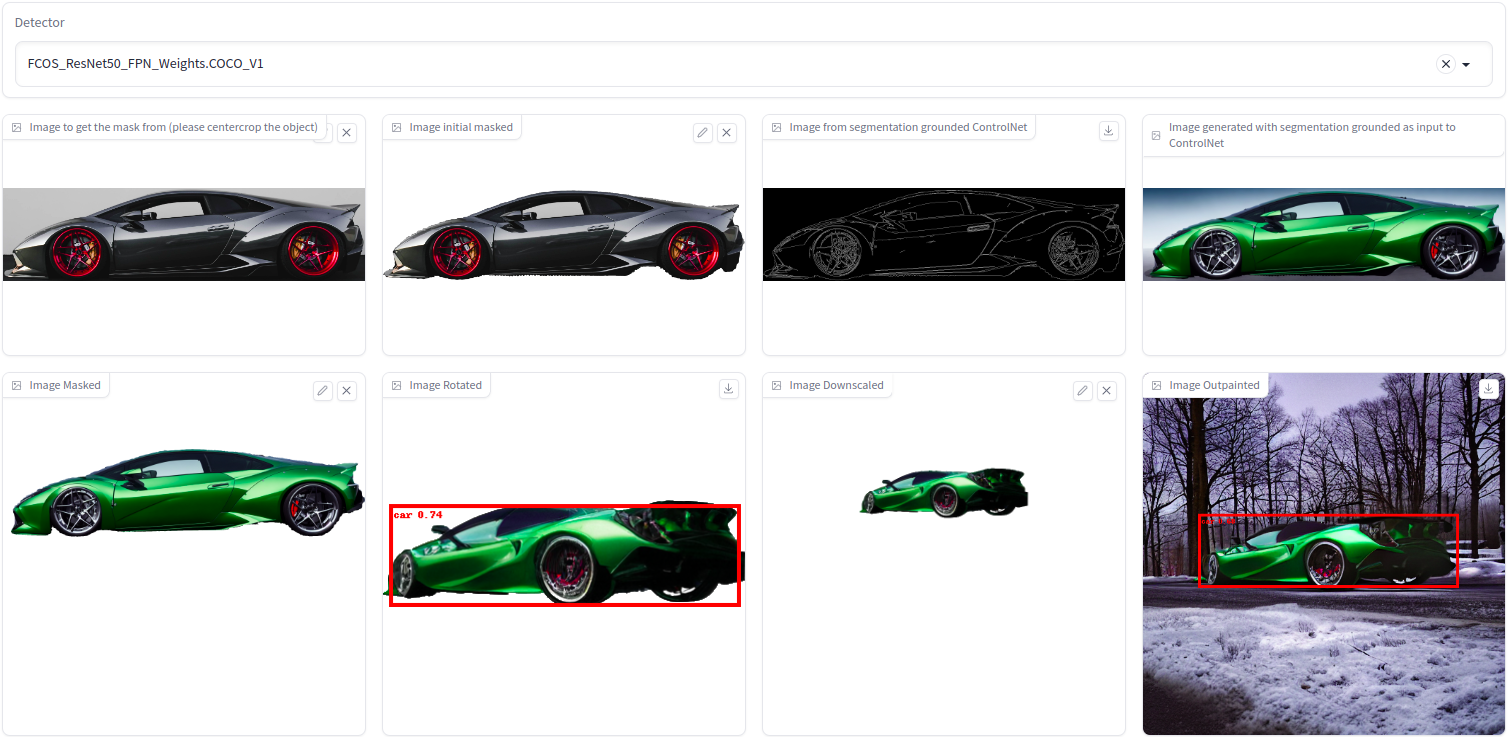}
  \end{subfigure}
  \caption{\textbf{Depiction of our SCROD pipeline for synthesizing objects with certain attributes}. The workflow is as follows:
  \textit{[Top, 1st from left]} We start with a real image of an object (in our experiments we always start with a side view of a car, but it can be chosen arbitrarily). 
  \textit{[Top, 2nd from left]}  We segment the object itself and remove the background, using SAM \cite{kirillov2023segany}. 
  \textit{[Top, 3rd from left]}  We extract Canny edge maps of the segmented object using a Canny edge detector.
  \textit{[Top, 4th from left]}  We invoke a ControlNet \cite{zhang2023adding} to synthesize a new image, conditioned on the edge map and a text prompt (in the example ``new green sports car''). The overall purpose of the top row is to control object attributes such as color (but also other types of textures or for instance, dirty-vs-clean could be controlled this way).
  \textit{[Bottom,1st from left]} The generated synthetic image is then again segmented, and the background is removed using SAM. 
  \textit{[Bottom, 2nd from left]} The segmented object is then rotated using Zero-1-to-3 \cite{liu2023zero1to3} by an externally specified rotation angle. Importantly, rotation does not require a 3D model of the object, just a segmented 2D view. Rotation can be around the axis left-right and top-bottom in principle (shown is only left-right). To improve image quality during the outpainting, we use Stable Diffusion x4 upscaler (SD x4) \cite{rombach2021highresolution}. Object detectors under investigation can be evaluated on these intermediate results for a plain color background (which reduces dependencies on background cues).
  \textit{[Bottom, 3rd from left]} The size and position of the object can be controlled by downsampling and translating the segmented object.
  \textit{[Bottom, 4th from left]} The white background can finally be ``outpainted'' by conditioning a text-to-image model on the object and a text prompt like ``sports car is driving on snowy street''. We propose to do the outpainting using the fine-tuned model using LoRA \cite{hu2022lora} and starting from the SD-v1.5 inpainting model as described in Section \ref{sec:pipeline}. The object detector under investigation can then be tested on the resulting image such that for instance systematic errors on realistic object backgrounds can be identified. 
  }
  \label{fig:pipeleine}
\end{figure*}

In contrast, we propose a multi-stage pipeline consisting of several generative models, where each model focuses on controlling specific properties of the generated objects and scenes. We focus on the properties such as \textit{object type, color, location, scale, orientation, and background} of a given object of category ``car''.  The whole pipeline is displayed step-by-step in Fig.~\ref{fig:pipeleine}. In the figure, we start with a real image of an object (in this case a side view of a black sports car). In the following, the runtime is reported for a batch size of $1$ on an A100 GPU.
\begin{enumerate}[noitemsep,topsep=0pt,parsep=0pt,partopsep=0pt]
    \item \textbf{Object type}: by using Segment Anything Model (SAM) \cite{kirillov2023segany} we segment objects of different car types automatically from the same fixed view (side view in our case). The inference time is $4$ seconds.
    \item \textbf{Object Color}: we extract Canny Edges of the segmented object and by using a ControlNet with Canny Edges conditioning \cite{zhang2023adding}, we then condition on the fine details of the automatically segmented object to create cars of different colors. Varying the color might introduce systematic errors for some object detectors as can be seen in the realistic setting in Fig.~\ref{fig:realistic_syserr_retina}. The inference time is $20$ seconds.
    \item \textbf{Orientation}: by using a model such as Zero-1-to-3 \cite{liu2023zero1to3}, that was fine-tuned from SD on ObjaVerse \cite{objaverse} to generate views of the object, conditioned on the desired angles around X, Y, and Z axes, we can rotate an object around the X axis (pointing up). In our pipeline, we additionally use Stable Diffusion x4 upscaler (SD x4) \cite{rombach2021highresolution} to improve the resolution of the generated images. 
    The inference time is $2$ seconds for the single view generation and $19$ seconds for upscaling.
    \item \textbf{Scale and Location}: we then downscale and change the location of the generated object by a different number of scaling factors, which does not require a separate model. Varying the scale and the location might introduce systematic errors for some object detectors as can be seen in the realistic setting in Fig.~\ref{fig:realistic_syserr_fcos}.

    \item \textbf{Background}: In the experiments of this paper, we focus mainly on a plain color background, which does not require a separate model, and reduces the effects of background cues on object detector behavior. However, in case we want to outpaint a naturally looking background instead of using a plain color, we require an outpainting model. For this we use LoRA fine-tuning \cite{hu2022lora} of the SD-v1.5 inpainting model as described in Section \ref{sec:outpainting_model}.
    In the case of using the outpainting model, the inference time is $5$ seconds.
\end{enumerate}

\begin{figure}[bt]
  \begin{minipage}{0.32\linewidth}
    \centering
    \includegraphics[width=\linewidth]{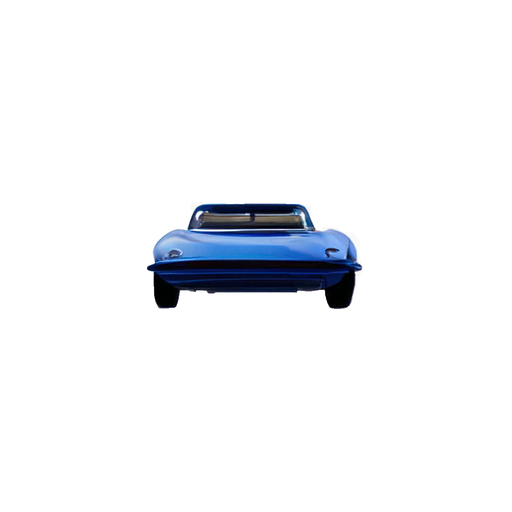}
  \end{minipage}%
  \begin{minipage}{0.32\linewidth}
    \centering
    \includegraphics[width=\linewidth]{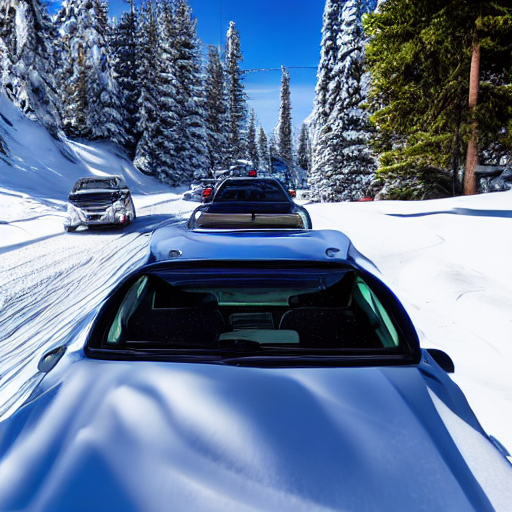}
  \end{minipage}
    \begin{minipage}{0.32\linewidth}
    \centering
    \includegraphics[width=\linewidth]{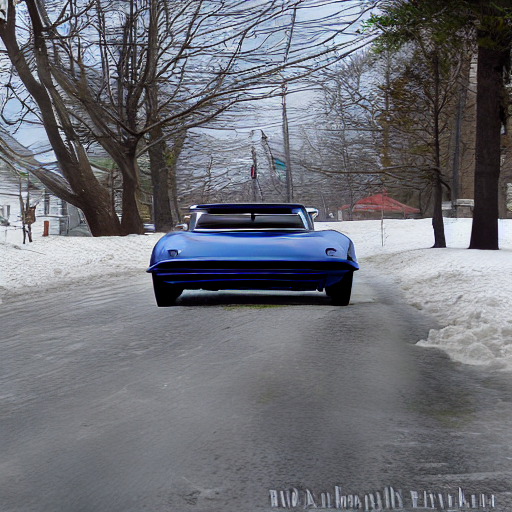}
  \end{minipage}
  \caption{\textbf{Object outpainting}: \textit{On the left} is the starting image of a synthetic car rotated by $90^\circ$. \textit{In the middle} is the image outpainted with the SD-v1.5 inpainting model, where we observe over-generation: instead of preserving the boundary of the object, the outpainting continues hallucinating the object outside of the given boundary. \textit{On the right} is the image outpainted with our proposed outpainting model, a LoRA fine-tuned \cite{hu2022lora} SD-v1.5 inpainting model (Section \ref{sec:outpainting_model}). Both outpaintings use the same seed and prompt \textit{``sedan is driving on snowy street''}.}
  \label{fig:outpainting_motivation}
\end{figure}

\subsection{Outpainting model}\label{sec:outpainting_model}
To date, no strong models for masked-object outpainting exist and inpainting models tend to \textit{over-generate}\footnote{We say an object is overgenerated if instead of preserving the boundary of the object during the outpainting continues hallucinating the object outside of the given boundary.} the object as can be seen in Fig.~\ref{fig:outpainting_motivation} (please note how the boundary of the ``car'' object in middle image is not preserved and the object is transformed into a bigger one). We use LoRA fine-tuning \cite{hu2022lora} of the SD-v1.5 inpainting model on the joint dataset of COCO \cite{lin2015microsoft} and BDD100k \cite{bdd100k}, with the following training details: \textit{number of training steps} is $15.000$ , \textit{LoRA rank} is $32$, \textit{batch size} is $8$, and \textit{learning rate} is $10^{-4}$. The dataset $(\mathcal{X}, \mathcal{C}, \hat{\mathcal{X}})$ with $N=19721$ samples has the structure as described below.
\vspace{-.1cm}
\begin{itemize}[noitemsep,topsep=0pt,parsep=0pt,partopsep=0pt]
    \item \textbf{Labels during LoRA fine-tuning:} Ground truth original images $\hat{\mathcal{X}} = (\hat{x}_i)_{i=1}^{N}$.
    \item \textbf{Inputs during LoRA fine-tuning:} Ground truth segmented objects $\mathcal{X} = (x_i)_{i=1}^{N}$ provided from COCO and BDD100k (note that only $8.000$ images from the whole BDD100k dataset have instance segmentations). During training, some objects in the segmentations masks are randomly dropped with probability $0.5$ if other objects cover at least $10\%$ of the area of the image. More precisely, $x_i = \hat{x}_i \cdot m_i$ for $i \in \{1, \dots, N\}$, where $m_i$ is a random binary mask obtained as described above. This is done to increase the diversity of the dataset by increasing the number of combinations of the masked objects shown on the image that are taken as an input. We consider objects from the following COCO classes as relevant to automated driving applications: \textit{car, person, truck, bus, traffic light, bicycle, motorcycle}; respectively from BDD100k: \textit{rider, car, truck, bus, train, motorcycle, bicycle}. 
    \item \textbf{Inputs during LoRA fine-tuning:} Captions $\mathcal{C} = (c_i)_{i=1}^{N}$, automatically generated using 
\textbf{BLIP-2}, OPT-2.7b model \cite{li2023blip2} $f_{\Theta,\mathrm{BLIP2}}$. That is, $c_i = f_{\Theta,\mathrm{BLIP2}}(\hat{x}_i)$. 
\end{itemize}

 Using this fine-tuned model, we can outpaint a realistic background, where the scene integrates the main object meaningfully, also generating shadows and reflections as can be seen in Fig.~\ref{fig:pipeleine}, \ref{fig:realistic_syserr_fcos} and \ref{fig:realistic_syserr_retina}.

\begin{table*}[bt]
\centering

\resizebox{\textwidth}{!}{%
\begin{tabular}{c c c c c | c c c c c} 
\hline
\multicolumn{5}{c|}{\textbf{Attributes}} & \multicolumn{5}{c}{\textbf{Error rates}} \\
\hline
scale & angle & O & BG & type & YOLOv5n %
& FCOS %
& RetinaNet2 %
& FasterRCNN2 %
& YOLOv5x6 %
\\ [0.5ex] 
\hline\hline
 - & $-90.0$ & black & blue & sports car & $98\%$ (airplane) & $90\%$ (mouse) & $96\%$ (mouse) & $94\%$ (kite) & $83\%$ (giraffe) \\ 

\cellcolor{blue!25} $6.0$ & \cellcolor{blue!25} $-50.0$ & \cellcolor{blue!25} - & \cellcolor{blue!25} grey & sports car & $70\%$ (airplane) & $50\%$ (kite) & $4\%$ (airplane) & \cellcolor{blue!25} $\mathbf{100\%}$ (airplane) & $0\%$ \\ 
$2.0$ &  $0.0$ & yellow & grey & sports car & $94\%$ (motorcycle) & $31\%$ (motorcycle) & $\mathbf{100\%}$ (motorcycle) & $0\%$ & $50\%$ (motorcycle) \\
$2.0$ &  $-90.0$ & pink & - & smart car & $\mathbf{100\%}$ (truck) & $\mathbf{100\%}$ (train) & $55\%$ (truck) & $19\%$ (kite) & $\mathbf{38\%}$ (truck) \\

$6.0$ &  $0.0$ & - & blue & sedan & $\mathbf{100\%}$ (airplane) & $8\%$ (clock) & $0\%$ & $0\%$ & $0\%$ \\
\hline
\multicolumn{5}{c|}{\textbf{Average Error Rate}} & $59\%$ & $30\%$ & $17\%$ & $27\%$ & $7\%$ \\
[1ex] 
\hline
\end{tabular}}
\caption{\textbf{Combinations of attributes scale, angle, object color, background color, and car type (column ``Attributes'' on the left) that have the highest error rate across seeds and marginalized attributes (if any) in the corresponding group for $3$ selected object detectors (column ``Error rates'' on the right)}. For each error rate, the class with the largest count of wrong predictions is displayed in parentheses for the object detector with the highest error rate. Here, we select the combinations of attributes that result in a significantly smaller error rate for the other $2$ object detectors, to highlight that these systematic errors are detector-specific. When we marginalize one of the attributes, we put "-" in the respective subcolumn of the column \textbf{``Attributes''}. For each object detector, we report additionally the respective Box MAP on COCO val2017 ($\text{B}_\text{M}$). In the last row, we report average error rates per object detector across all subgroups. It shows that object detectors can deal well with our synthetic images.} 
\label{tab:highest_error_rate}
\end{table*}

\section{Evaluation}\label{sec:evaluation}

We make use of the SCROD pipeline described in Section \ref{sec:pipeline} to test the $3$ best object detectors from torchvision \cite{torchvision2016} according to the Box MAP on COCO val2017 ($\text{B}_\text{M}$):
\begin{itemize}[noitemsep,topsep=0pt,parsep=0pt,partopsep=0pt]
  \item \mbox{FasterRCNN\_ResNet50\_FPN\_V2 (\textbf{FasterRCNN2})}\cite{li2021benchmarking}
  \item \mbox{RetinaNet\_ResNet50\_FPN\_V2 (\textbf{RetinaNet2})} \cite{zhang2020bridging}
  \item \mbox{FCOS\_ResNet50\_FPN (\textbf{FCOS})} \cite{tian2019fcos}
\end{itemize}
\subsection{Color background}\label{sec:color_background}
\begin{figure*}[tb]
  \centering
  \begin{subfigure}{0.48\textwidth}
    \centering
    \includegraphics[width=\linewidth]{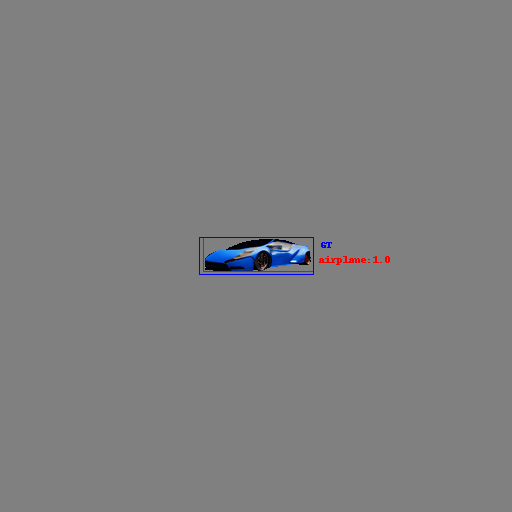}
  \end{subfigure}
  \quad
  \begin{subfigure}{0.48\textwidth}
    \centering
    \includegraphics[width=\linewidth]{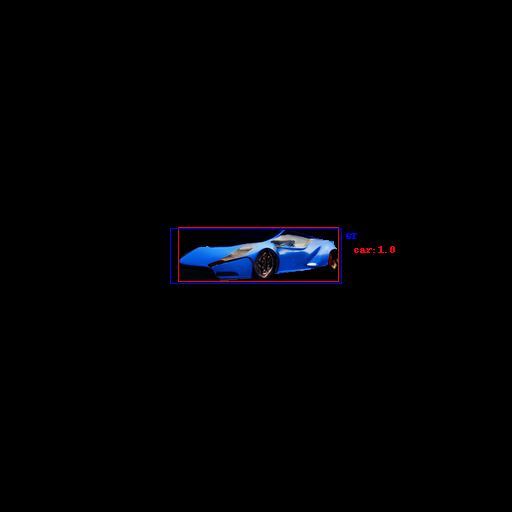}
  \end{subfigure}
  \caption{\textbf{Systematic error for the object detector ``FasterRCNN2'' motivated by the selected entry from the Tab.~\ref{tab:highest_error_rate} (highlighted in blue)}. \textit{On the left}, is the image, which is detected as the class \textbf{airplane}. This happens for the same attributes and same object detector across all $16$ seeds when randomizing single view generation with Zero-1-to-3. Attributes used during the generation are the same as in the selected entry in the table. 
  \textit{On the right}, is the image for the same seed, but where the downscaling factor is decreased from $6.0$ to $4.0$ and the background color is changed to black.}
  \label{fig:highest_error_rate_fasterRCNN}
\end{figure*}

We start by varying the background colors and showing the systematic errors of object detectors when evaluated on the images generated with our pipeline and $11$ different background colors over $16$ seeds, when randomizing single view generation with Zero-1-to-3, for the generated object. In Tab.~\ref{tab:highest_error_rate} we show some of the combinations of attributes for each of the object detectors with the highest error rate. An example of a systematic error and how minor changes to the underlying attributes can remove it is shown in Fig.~\ref{fig:highest_error_rate_fasterRCNN}. 

By using our SCROD pipeline, we can find consistent and detector-specific systematic errors - that they occur predominantly for a single detector indicates that they are not caused by the generation process itself (for example, for the systematic error described in the first row of Tab.~\ref{tab:highest_error_rate}, images show cars and not airplanes, as is shown in Fig.~\ref{fig:highest_error_rate_fasterRCNN}). Also, we can highlight that systematic errors occur only in very specific subgroups and minor changes remove them. Lastly, the average error rates of detectors across all subgroups reported in the last row of Tab.~\ref{tab:highest_error_rate} show that the object detectors can deal well with our synthetic images. 

\subsection{Realistic background}
\begin{figure*}[tb]
  \centering
  \begin{subfigure}{0.45\textwidth}
    \centering
    \includegraphics[width=\linewidth]{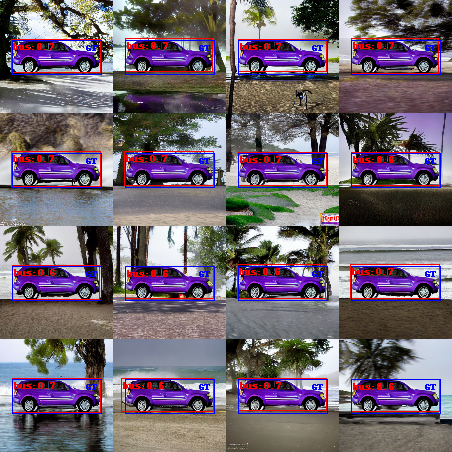}
    \caption{Systematic error into the class \textbf{bus} for the prompt \textit{``purple SUV is driving on beach, foggy, idyllic, lush detail''}. Here, the image is downscaled by a factor of $4.5$ before inputting it into the object detector and the rotation angle is $10^{\circ}$. \newline }
  \end{subfigure}
  \quad
  \begin{subfigure}{0.45\textwidth}
    \centering
    \includegraphics[width=\linewidth]{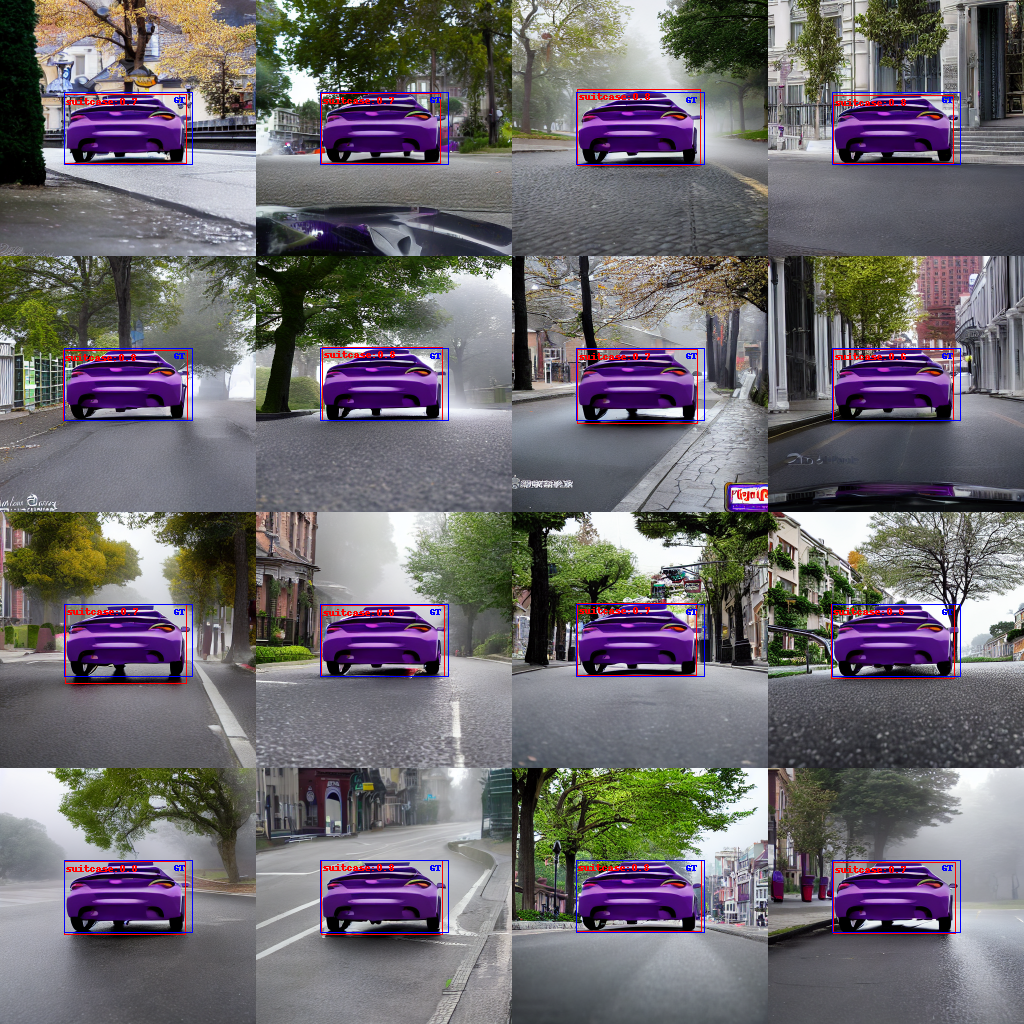}
    \caption{Systematic error into the class \textbf{suitcase} for the prompt \textit{``purple coupe car is driving on street, foggy, in the morning, bright, stunning environment, sharp focus''}. Here, the image is downscaled by a factor of $2.0$ before inputting it into the object detector and the rotation angle is $-80^{\circ}$.}
  \end{subfigure}
    \begin{subfigure}{0.45\textwidth}
    \centering
    \includegraphics[width=\linewidth]{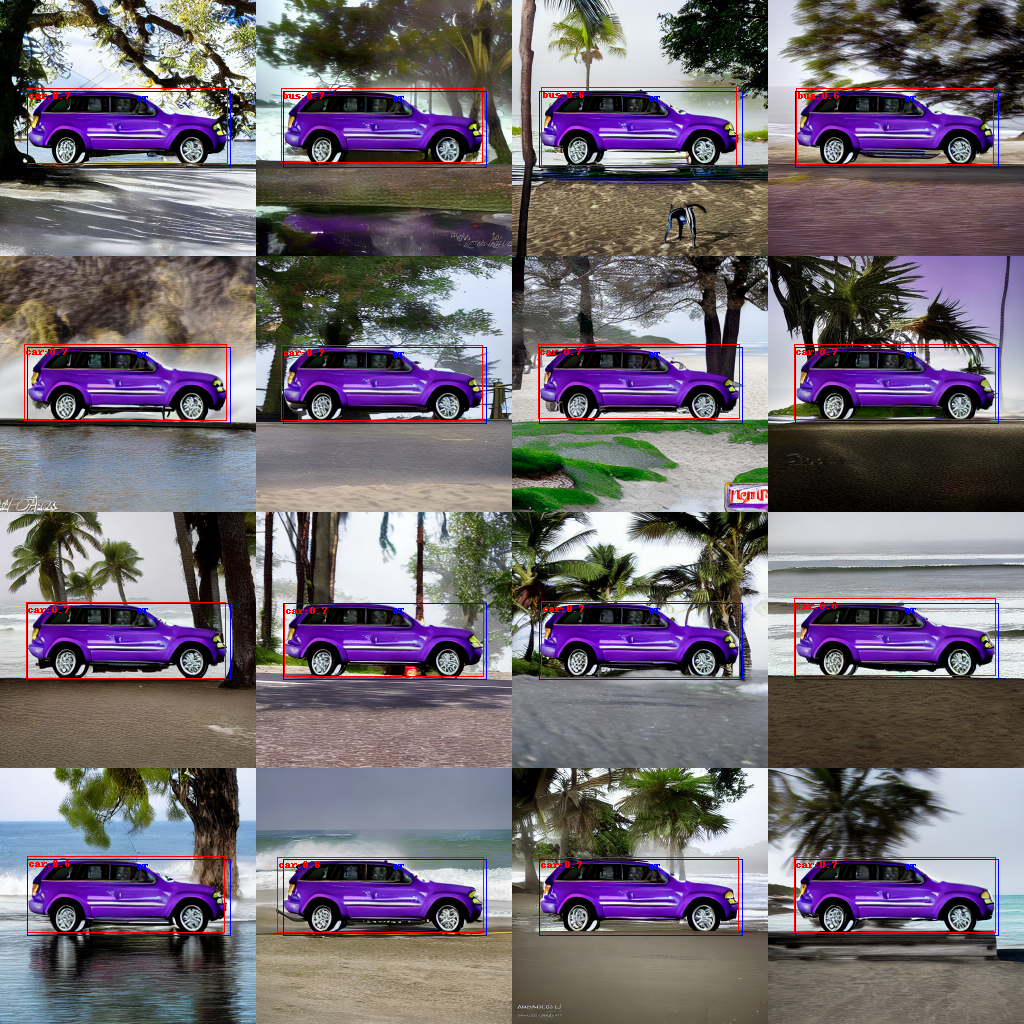}
    \caption{Changing the prediction from the class \textbf{bus} to the correct class \textbf{car} for the prompt \textit{``purple SUV is driving on beach, foggy, , idyllic, lush detail''} by changing the downscaling factor from $4.5$ to $2.0$ before inputting it into the object detector. Here the rotation angle is $10^{\circ}$. \newline }
  \end{subfigure}
  \quad
  \begin{subfigure}{0.45\textwidth}
    \centering
    \includegraphics[width=\linewidth]{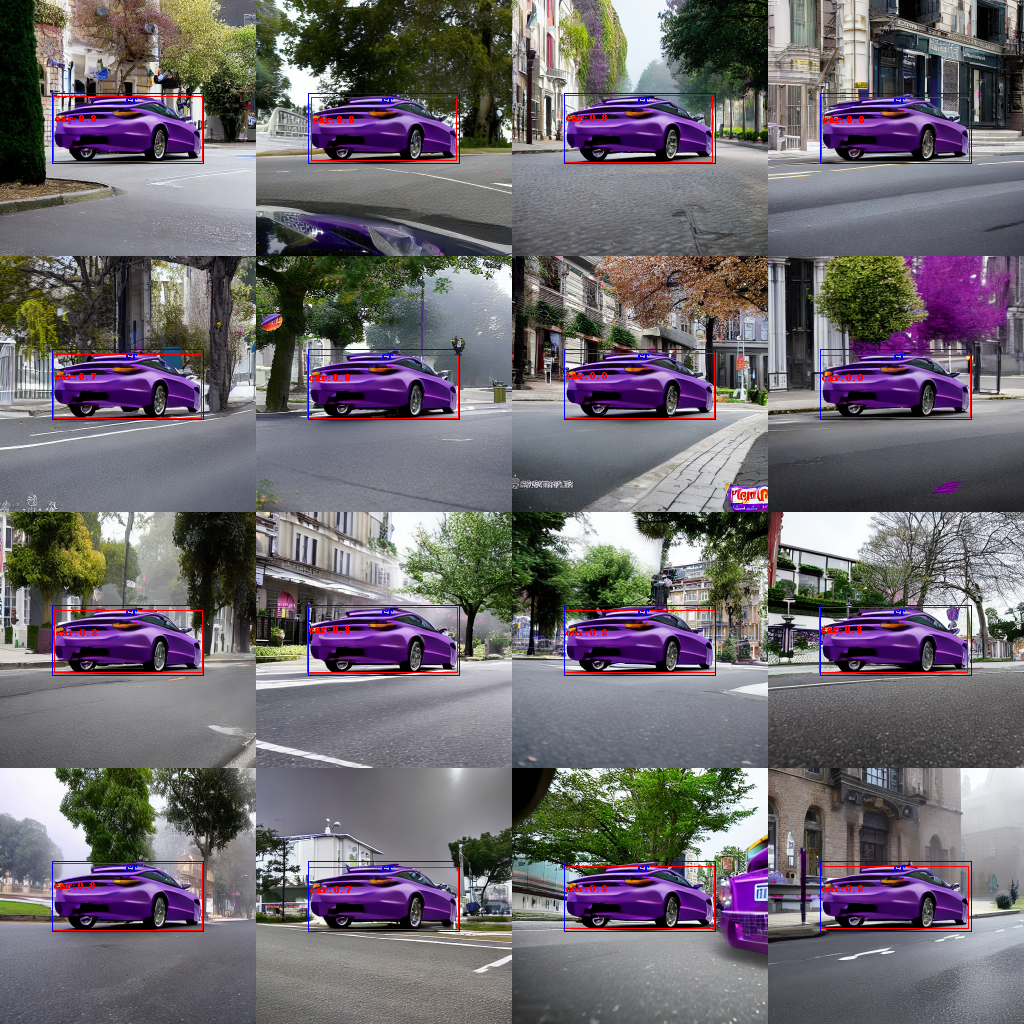}
    \caption{Changing the prediction from the class \textbf{suitcase} to the correct class \textbf{car} for the prompt \textit{``purple coupe car is driving on street, foggy, in the morning, bright, stunning environment, sharp focus''} by changing the rotation angle from $-80^{\circ}$ to $-70^{\circ}$. Here, the image is downscaled by a factor of $2.0$ before inputting it into the object detector.}
  \end{subfigure}
  \caption{\textbf{Subfigures (a) and (b) show detector-specific systematic errors for the object detector ``FCOS''}. Subfigures (c) and (d) show that minor changes in object pose and scale result in correct predictions, which justifies that fine-granular control over image synthesis is required for identifying systematic errors. While all of the objects in (a) and (b) are detected as a wrong class with the object detector \textbf{``FCOS''} (error rate is $100\%$),\textbf{``RetinaNet2''} has error rates $13\%$ (a), $25\%$ (b), and \textbf{``FasterRCNN2''} - $0\%$ in both cases.}
  \label{fig:realistic_syserr_fcos}
\end{figure*}

\begin{figure*}[tb]
  \centering
  \begin{subfigure}{0.45\textwidth}
    \centering
    \includegraphics[width=\linewidth]{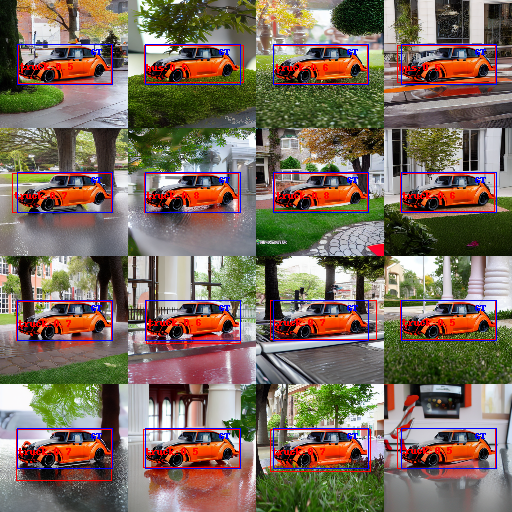}
    \caption{Systematic error into the classes \textbf{truck} and \textbf{bus} for the prompt \textit{``orange smart car is driving on lawn, rainy, in the morning, bright, lush vegetation, HQ''}. Here, the image is downscaled by a factor of $4.0$ before inputting it into the object detector and the rotation angle is $-30^{\circ}$.}
  \end{subfigure}
  \quad
  \begin{subfigure}{0.45\textwidth}
    \centering
    \includegraphics[width=\linewidth]{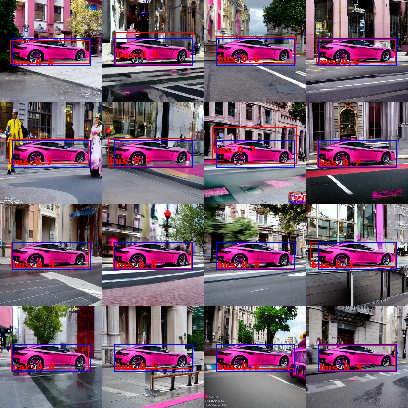}
    \caption{Systematic error into the class \textbf{bus} for the prompt \textit{``pink coupe car is driving in city, thundery, in the morning, bright, landscape, HQ''}. Here, the image is downscaled by a factor of $5.0$ before inputting it into the object detector and the rotation angle is $-30^{\circ}$.}
  \end{subfigure}
    \begin{subfigure}{0.45\textwidth}
    \centering
    \includegraphics[width=\linewidth]{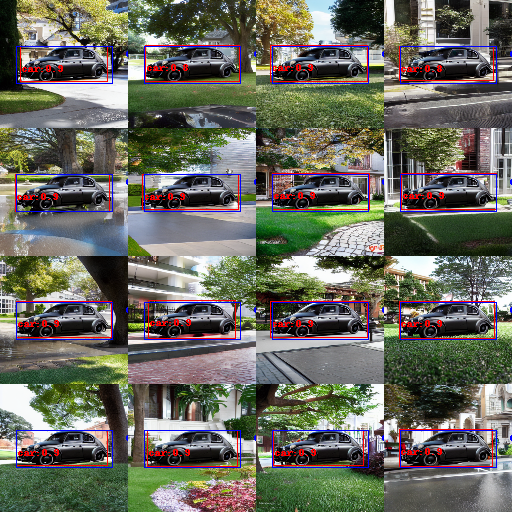}
    \caption{Changing the prediction from the classes \textbf{truck} and \textbf{bus} to the correct class \textbf{car} for the prompt \textit{``black smart car is driving on lawn, rainy, in the morning, bright, lush vegetation, HQ''} by changing the color from ``orange'' to ``black''. Here, the image is downscaled by a factor of $4.0$ before inputting it into the object detector and the rotation angle is $-30^{\circ}$.}
  \end{subfigure}
  \quad
  \begin{subfigure}{0.45\textwidth}
    \centering
    \includegraphics[width=\linewidth]{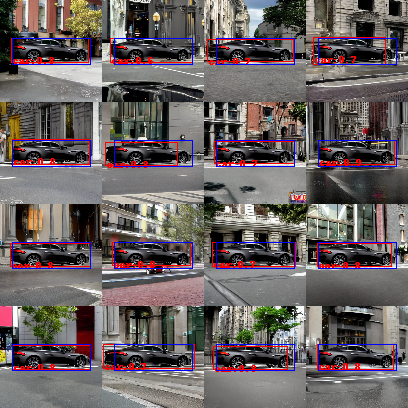}
    \caption{Changing the prediction from the class \textbf{bus} to the correct class \textbf{car} for the prompt \textit{``black coupe car is driving in city, thundery, in the morning, bright, landscape, HQ''} by changing the color from ``pink'' to ``black''. Here, the image is downscaled by a factor of $5.0$ before inputting it into the object detector and the rotation angle is $-30^{\circ}$.}
  \end{subfigure}
  \caption{\textbf{Subfigures (a) and (b) show detector-specific systematic errors for the object detectors ``RetinaNet2'' and ``FCOS''}. Subfigures (c) and (d) show that minor changes in object color result in correct predictions, which justifies that fine-granular control over image synthesis is required for identifying systematic errors. While all of the objects in (a) and all but $1$ in (b) are detected as a wrong class with the object detector \textbf{``RetinaNet2''} (error rate is $100\%$ and $94\%$ respectively), and all are detected as a wrong class with the object detector \textbf{``FCOS''}, detector \textbf{``FasterRCNN2''} has error rate $0\%$ in both cases.}
  \label{fig:realistic_syserr_retina}
\end{figure*}

Moreover, we show systematic errors with a more complex (outpainted) background. Here we fix one seed for the generated object and vary the seed for the outpainted background. As can be seen from the reflections and shadows on images in Fig.~\ref{fig:realistic_syserr_fcos} and \ref{fig:realistic_syserr_retina}, our proposed pipeline together with the custom model for outpainting allows us to generate scenes that look more natural and cannot be achieved by simply pasting an object into an existing background. 

A more detailed analysis requires varying all possible single attributes and observing, which other combinations of the attributes can lead to the same systematic error, and which smallest changes of them can remove the systematic error. One example of such observation can be seen in Fig.~\ref{fig:realistic_syserr_fcos} (c) and (d). From it, we can see that systematic errors are brittle and occur only under very controlled conditions. 

By using our SCROD pipeline, we can easily identify such systematic errors of the object detector ``FCOS'' as a dependance on the scale of the object or on the angle of rotation of the object not only on the plain color background but also on a natural background, when outpainted using our LoRA fine-tuned model. This can be seen in Fig.~\ref{fig:realistic_syserr_fcos}. Here, we use $16$ seeds when randomizing the outpainting.
\begin{enumerate}[noitemsep,topsep=0pt,parsep=0pt,partopsep=0pt]
    \item In Fig.~\ref{fig:realistic_syserr_fcos}, (a) and (c), %
    a purple SUV on the background outpainted using the prompt \textit{``purple SUV is driving on beach, foggy, idyllic, lush detail.''} is incorrectly detected as \textbf{bus} for all $16$ seeds in (a). By changing the downscaling factor from $4.5$ to $2.0$, for $13$ out of $16$ seeds, objects are correctly detected as \textbf{car} (c).
    \item In Fig.~\ref{fig:realistic_syserr_fcos}, (b) and (d), %
    a purple coupe car on the background outpainted using the prompt \textit{``purple coupe car is driving on street, foggy, in the morning, bright, stunning environment, sharp focus''} is incorrectly detected as \textbf{suitcase} for all $16$ seeds in (b). By changing the rotation angle from $-80^\circ$ to $-70^\circ$, for $16$ out of $16$ seeds, objects are correctly detected as \textbf{car} (d).
\end{enumerate} 
Similarly, as displayed in Fig.~\ref{fig:realistic_syserr_retina}, we can easily identify another systematic error of the object detector ``RetinaNet2'' as a dependence on the color of the object.

\section{Conclusions}
\textbf{Summary.} In this paper, we introduce a novel pipeline called SCROD, designed to automatically identify systematic errors in object detectors applied to synthetic street scenes. Our pipeline incorporates a custom outpainting model%
, enabling comprehensive error analysis in both plain color and natural backgrounds. We show the brittleness of even state-of-the-art object detectors from torchvision \cite{torchvision2016}, highlighting the need for improved evaluation protocols.

\textbf{Limitations.} 
While SCROD provides extensive control over relevant attributes, the quality of control is still subject to the existing models' limitations. Additionally, our current implementation is object-centric and checks if a single object 
is detected correctly, potentially leaving out certain testing scenarios such as two objects occluding each other.

\textbf{Outlook.} Our SCROD pipeline is modular and building blocks such as ControlNet, Zero-1-to-3, or Outpainting can easily be replaced once better alternatives become available. So we expect the quality of our object detector testing pipeline SCROD will improve in the future. We leave it for future research to extend SCROD to handle multiple objects and occlusions within the scene. Additionally, developing more efficient non-brute-force search procedures is a promising direction. %

\clearpage
\clearpage

{\small
\bibliographystyle{ieee_fullname}

}

\end{document}